\documentclass[10pt,twocolumn,letterpaper]{article}

\usepackage{cvpr}
\usepackage{times}
\usepackage{epsfig}
\usepackage{graphicx}
\usepackage{amsmath}
\usepackage{amssymb}
\usepackage{makecell}
\usepackage{caption}
\DeclareMathOperator*{\argmin}{arg\,min}

\usepackage[pagebackref=true,breaklinks=true,letterpaper=true,colorlinks,bookmarks=false]{hyperref}

 \cvprfinalcopy 

\ifcvprfinal\fi
\begin{document}

\title{Learning to Generate Dense Point Clouds with Textures on Multiple Categories}

\author{Tao Hu, Geng Lin, Zhizhong Han, Matthias Zwicker\\
Department of Computer Science, University of Maryland, College Park\\
{\tt\small taohu@cs.umd.edu, geng@cs.umd.edu, h312h@umd.edu, zwicker@cs.umd.edu}
}

\maketitle

\begin{abstract}
3D reconstruction from images is a core problem in computer vision. With recent advances in deep learning, it has become possible to recover plausible 3D shapes even from single RGB images for the first time. However, obtaining detailed geometry and texture for objects with arbitrary topology remains challenging. In this paper, we propose a novel approach for reconstructing point clouds from RGB images. Unlike other methods, we can recover dense point clouds with hundreds of thousands of points, and we also include RGB textures. In addition, we train our model on multiple categories which leads to superior generalization to unseen categories compared to previous techniques. We achieve this using a two-stage approach, where we first infer an object coordinate map from the input RGB image, and then obtain the final point cloud using a reprojection and completion step. We show results on standard benchmarks that demonstrate the advantages of our technique. Code is available at \href{https://github.com/TaoHuUMD/3D-Reconstruction}{https://github.com/TaoHuUMD/3D-Reconstruction}
\end{abstract}

\section{Introduction}

3D reconstruction from single RGB images has been a longstanding challenge in computer vision. While recent progress with deep learning-based techniques and large shape or image databases has been significant, the reconstruction of detailed geometry and texture for a large variety of object categories with arbitrary topology remains challenging. Point clouds have emerged as one of the most popular representations to tackle this challenge because of a number of distinct advantages: unlike meshes they can easily represent arbitrary topology, unlike 3D voxel grids they do not suffer from cubic complexity, and unlike implicit functions they can reconstruct shapes using a single evaluation of a neural network. In addition, it is straightforward to represent surface textures with point clouds by storing per-point RGB values.

In this paper, we present a novel method to reconstruct 3D point clouds from single RGB images, including the optional recovery of per-point RGB texture. In addition, our approach can be trained on multiple categories. The key idea of our method is to solve the problem using a two-stage approach, where both stages can be implemented using powerful 2D image-to-image translation networks: in the first stage, we recover an object coordinate map from the input RGB image. This is similar to a depth image, but it corresponds to a point cloud in object-centric coordinates that is independent of camera pose. In the second stage, we reproject the object space point cloud into depth images from eight fixed viewpoints in image space, and perform depth map completion. We can then trivially fuse all completed object space depth maps into a final 3D reconstruction, without requiring a separate alignment stage, for example using  iterative closest point algorithm (ICP) \cite{icp_alg}. Since all networks are based on 2D convolutions, it is straightforward to achieve high resolution reconstructions with this approach. Texture reconstruction uses the same pipeline, but operating on RGB images instead of object space depth maps.

We train our approach on a multi-category dataset and show that our object-centric, two-stage approach leads to better generalization than competing techniques. In addition, recovering object space point clouds allows us to avoid a separate camera pose estimation step. In summary, our main contributions are as follows:


\begin{itemize}
	\item A strategy to generate 3D shapes from single RGB images in a two-stage approach, by first recovering object coordinate images as an intermediate representation, and then performing reprojection, depth map completion, and a final trivial fusion step in object space.
	\item The first work to train a single network to reconstruct point clouds with RGB textures on multiple categories.
	\item More accurate reconstruction results than previous methods on both seen and unseen categories from ShapeNet \cite{shapenet} or Pix3D \cite{pix3d} datasets.
\end{itemize}

\begin{figure*}
	\begin{center}
		\includegraphics[width=\linewidth]{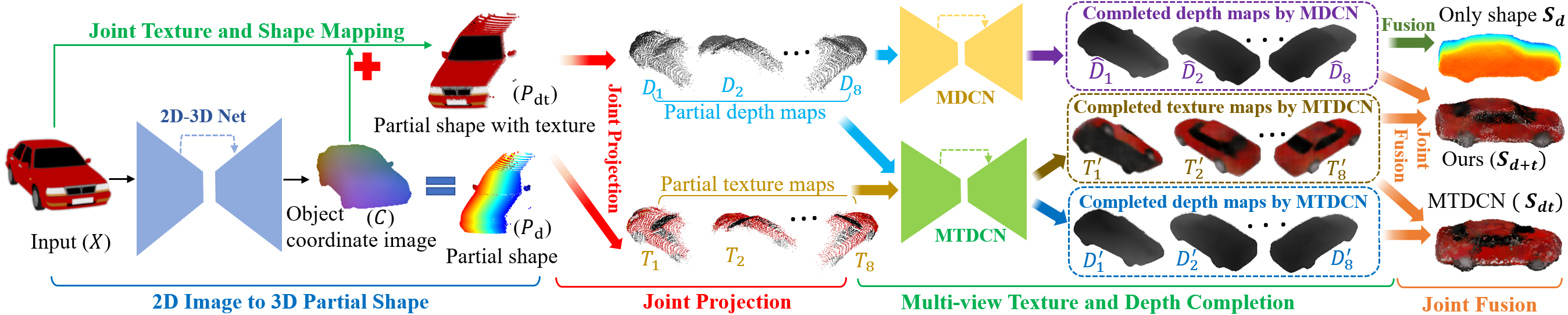}
	\end{center}
	\vspace{-0.13in}
	\caption{Approach overview. An image $X$ is passed through a 2D-3D network to reconstruct the visible parts of the object, represented by an object coordinate image $C$. $X$ and $C$ represent the texture and 3D coordinates of a shape respectively, which yield a partial shape with texture $P_{dt}$ when combined by a \textit{Joint Texture and Shape Mapping} operator. Next, by \textit{Joint Projection}, $P_{dt}$ is jointly projected from 8 fixed viewpoints into 8 pairs of partial depth maps and textures maps, which are translated to completed maps by the Multi-view Texture-Depth Completion Net (MTDCN) that jointly completes texture and depth maps. Alternatively, Multi-view Depth Completion Net (MDCN) only completes the depth maps. Finally, the \textit{Joint Fusion} operator fuses the completed multiple texture and depth maps into completed point clouds. }
	\label{fig:overview}
		\vspace{-0.1in}
\end{figure*}

\section{Related Work}
Our method is mainly related to single image 3D reconstruction and shape completion. We briefly review previous works in these two aspects.

\noindent \textbf{{Single image 3D reconstruction}}.
Along with the development of deep learning techniques, single image 3D reconstruction has made a huge progress. Because of the regularity, early works mainly learned to reconstruct voxel grids from 3D supervision~\cite{r2n2} or 2D supervision~\cite{Tatarchenko2015Multiview3M} using differentiable renderers~\cite{ptn,mvcTulsiani18}. However, these methods can only reconstruct shapes at low resolution, such as 32 or 64, due to the cubic complexity of voxel grids. Although various strategies~\cite{DBLP:conf/3dim/HaneTM17,DBLP:conf/iccv/TatarchenkoDB17} were  proposed to increase the resolution, these methods were too complex to follow. Mesh based methods~\cite{WangZLFLJ18,liu2019softras} are also alternatives to increase the resolution. However, these methods are still hard to handle arbitrary topology, since the vertices topology of reconstructed shapes mainly inherits from the template. Point clouds based methods~\cite{ref_cd,pointnetpp,depth_zeng,lmnet} provides another direction for single image 3D reconstruction. However, these methods also have a bottleneck of low resolution, which makes it hard to reveal more geometry details.

Besides low resolution, lack of texture is another issue which significantly affects the realism of the generated shapes. Current methods aim to map the texture from single images to reconstructed shapes either represented by mesh templates~\cite{bird} or point clouds in a form of object coordinate maps~\cite{NIPS2019_Srinath}. Although these methods have shown promising results in some specific shape classes, they usually can only work in  category-specific reconstruction. In addition, the texture prediction pipeline of ~\cite{bird} sampling pixels from input images directly work on symmetric object with a good viewpoint. Though some other methods (e.g. \cite{Zhu2018VisualON, Tatarchenko2015Multiview3M}) predict nice novel RGB views by view synthesis, they can only work on category-specific reconstruction.


Different from all these methods, our method can jointly learn to reconstruct high resolution geometry and texture by a two-stage reconstruction and taking  object coordinate maps (also called NOCS map in \cite{Wang2019NormalizedOC,NIPS2019_Srinath}) as intermediate representation. Different from previous methods \cite{mit_unseen,depth_zeng} which use depth maps as intermediate representation and require camera pose information in their pipelines, our method does not require camera pose information. 

\noindent \textbf{{Shape completion}}. Shape completion is to infer the whole 3D geometry from partial observations. Different methods use volumetric grids \cite{epn3d} or point clouds \cite{ref_pcn,folding,fc_Achlioptas2018LearningRA} as shape representation for completion task. Points-based methods are mainly based on encoder and decoder structure which employs PointNet architecture~\cite{pointnet} as backbones. Although these works have shown nice completed shapes, they are limited to low resolution. To resolve this issue, Hu et al.~\cite{mvcn} introduced  Render4Completion to cast the 3D shape completion problem into multiple 2D view completion, which demonstrates promising potential on high resolution shape completion. Our method follows this direction, however, we can not only learn geometry but also texture, which makes our method much different.

\section{Approach}


Most 3D point cloud reconstruction methods \cite{lmnet,r2n2,psgn} solely focus on generating 3D shapes $\{P_i=[x_i, y_i, z_i]\}$ from input RGB images $X \in \mathbb{R}^{H \times W \times 3}$, where $H \times W$ is the image resolution and $[x_i, y_i, z_i]$ are 3D coordinates. Recovering the texture besides 3D coordinates is a more challenging task, which requires learning a mapping from $\mathbb{R}^{H \times W \times 3}$ to $\{P_i=[x_i, y_i, z_i, r_i, g_i, b_i]\}$, where $[r_i, g_i, b_i]$ are RGB values. 

We propose a method to generate high resolution 3D predictions and recover textures from RGB images. At a high level, we decompose the reconstruction problem into two less challenging tasks: first, transforming 2D images to 3D partial shapes that correspond to the observed parts of the target object, and second, completing the unseen parts of the 3D object. We use object coordinate images to represent partial 3D shapes, and multiple depth and RGB views to represent completed 3D shapes. 

As shown in Fig.~\ref{fig:overview}, our pipeline consists of four sub-modules: (1) 2D-3D Net, an image translation network which translates an RGB image $X$ to a partial shape $P_d$ (represented by object coordinate image $C$); (2) the Joint Projection module, which first jointly maps the partial shape $P_d$ with texture $X$ to generate $P_{dt}$, a partial shape mapped with texture, and then jointly project $P_{dt}$ into 8 pairs of partial depth $[D_1, \dots, D_8]$ and texture views $[T_1,\dots,T_8]$ from 8 fixed viewpoints (the 8 vertices of a cube); (3) the multi-view texture and depth completion module, which consists of two networks: Multi-view Texture-Depth Completion Net (MTDCN), which generates completed texture maps $[T'_1,\dots,T'_8]$ and depth maps $[D'_1,\dots,D'_8]$ by jointly completing partial texture and depth maps, and as an alternative, Multi-view Texture-Depth Completion Net (MDCN), which only completes depth maps and generates more accurate results $[\hat{D}_1,\dots,\hat{D}_8]$; (4) the Joint Fusion module, which jointly fuses the completed depth and texture views into completed 3D shape with textures, like $S_{d+t}$ and $S_{dt}$.

\subsection{2D RGB Image to Partial Shapes}

We propose to use 3-channel object coordinate images to represent partial shapes. Each pixel on the object coordinate image represents a 3D point, where its $(r, g, b)$ value corresponds to the point's location $(x,y,z)$. An object coordinate image is aligned with the input image, as shown in Figure \ref{fig:overview}, and in our pipeline, it represents the visible parts of the target 3D object. With this image-based 3D representation, we formulate the 2D-to-3D transformation as an image-to-image translation problem, and propose a 2D-3D Net to perform the translation based on the U-Net \cite{unet} architecture as in \cite{pix2pix2016}.

Unlike the depth map representation used in \cite{mit_unseen} and \cite{depth_zeng}, which requires camera pose information for back-projection, the 3-channel object coordinate image can represent a 3D shape independently. Note that our network infers the camera pose of the input RGB image so that the generated partial shape is aligned with ground truth 3D shape.

\subsection{Partial Shapes to Multiple Views}

\begin{figure}
	\begin{center}
		\includegraphics[width=\linewidth]{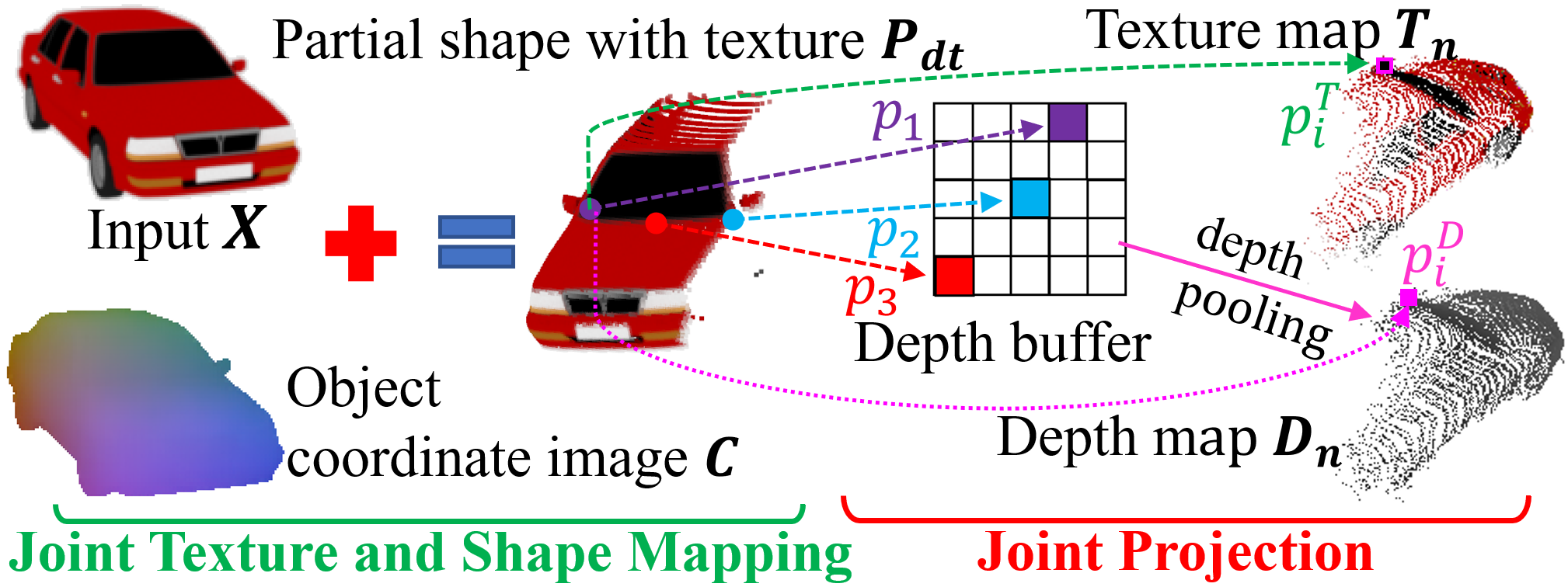}
	\end{center}
	\vspace{-0.13in}
	\caption{Joint Projection.}
	\label{fig:pooling}
	\vspace{-0.15in}
\end{figure}

In this module, we transform the input RGB image $X$ and the predicted object coordinate image $C$ to a partial shape mapped with texture, $P_{dt}$, which is then rendered from 8 fixed viewpoints to generate depth maps and texture maps. The process is illustrated in Fig. \ref{fig:pooling}.

\noindent \textbf{Joint Texture and Shape Mapping}. The input RGB image $X$ is aligned with the generated object coordinate image $C$. An equivalent partial point cloud $P_{dt}$ can be obtained by taking 3D coordinates from $C$ and texture from $X$. 


We denote a pixel on $X$ as $p^X_i = [{u}^X_i, {v}^X_i, {r}^X_i, {g}^X_i, {b}^X_i]$, where ${u}^X_i$ and ${v}^X_i$ are pixel coordinates, and similarly, a point on $C$ as $p^C_i = [{u}^C_i, {v}^C_i, {x}^C_i, {y}^C_i, {z}^C_i]$. Given $p^X_i$ and $p^C_i$  appearing at the same location, which means ${u}^X_i = {u}^C_i$ and ${v}^X_i = {v}^C_i$,  then $p^X_i$ and $p^C_i$ can be projected into 3D coordinates as ${P}_i = [x_i, y_i, z_i, r_i, g_i, b_i]$ on partial shape $P_{dt}$, where $r_i, g_i, b_i$ are RGB channels and $x_i={x}^C_i, y_i={y}^C_i, z_i={z}^C_i, r_i={r}^X_i, g_i={g}^X_i, b_i={b}^X_i$. 

\noindent \textbf{Joint Projection}. We render multiple depth maps $D=\{D_1,\dots,D_8\}$ and texture maps $T=\{T_1,\dots,T_8\}$ from 8 fixed viewpoints $V=\{V_1,\dots,V_8\}$ of the partial shape $P_{dt}$, where $D_n \in \mathbb{R}^{H \times W}$,  $T_n \in \mathbb{R}^{H \times W \times 3} , n \in [1,8]$.

Given $n$, we denote a point on depth map $D_n$ as $p^D_i = [{u}^D_i, {v}^D_i, {d}^D_i]$ where ${u}^D_i$ and ${v}^D_i$ are pixel coordinates and ${d}^D_i$ is the depth value. Similarly, a point on $T_n$ is $p^T_i = [{u}^T_i, {v}^T_i, {r}^T_i, {g}^T_i, {b}^T_i]$, where ${r}^T_i, {g}^T_i, {b}^T_i$ are RGB values. Then, we transform each 3D point $P_i$ on the partial shape $P_{dt}$ into a pixel $p'_i=[{u}'_i, {v}'_i, {d}'_i]$ on depth map $D_n$ by 
\begin{equation}
\vspace{-0.04in}
{p}'_i=K(\Re_n {P}_i + \tau_n)  \quad  \forall i,
\label{eq:project}
\end{equation}
where $K$ is the intrinsic camera matrix, $\Re_n$ and $\tau_n$ are the rotation matrix and translation vector of view $V_n$. Note that Eq. (\ref{eq:project}) only projects the 3D coordinates of $P_i$.

However, different points on $P_{dt}$ may be projected to the same location $[u,v]$ on the depth map $D_n$. For example, in Fig. \ref{fig:pooling}, ${p_1}=[{u},{v}, {d}_1], {p_2}=[{u},{v}, {d}_2], {p_3}=[{u},{v}, {d}_3]$ are projected to the same pixel ${p^D_i} = [{u}^D_i, {v}^D_i, {d}^D_i]$ on $D_n$, where $u^D_i=u,v^D_i=v$. The corresponding point on the texture map $T_n$ is $p^T_i = [{u}^T_i, {v}^T_i, {r}^T_i, {g}^T_i, {b}^T_i]$ where $u^T_i=u,v^T_i=v$.

To alleviate this collision effect, We implement a pseudo-rendering technique similar to ~\cite{hu20193d,lin2018learning}. Specifically, for each point on $P_{dt}$, a depth buffer with a size of $U \times U$ is used to store multiple depth values corresponding to the same pixel. Then we implement a depth-pooling operator with stride $U \times U$ to select the minimum depth value. We set $U=5$ in our experiments. In depth-pooling, we store the indices of pooling ($j$) and select the closest point from the view point $V_n$ among $\{{p_1},{p_2},{p_3}\}$. For example, in Fig.~\ref{fig:pooling}, pooling index $j=1$, the selected point is ${p_1}$, and the corresponding point on $P_{dt}$ is $P_1$. In this case, we copy the texture values from $P_1$ to $p^T_i$.

\subsection{Multi-view Texture and Depth Completion}
In our pipeline, a full shape is represented by depth images from multiple views, which are processed by CNNs to generate high resolution 3D shapes as mentioned in \cite{lin2018learning, mvcn}.

\begin{figure}[t]
	\begin{center}
		\includegraphics[width=1.02\linewidth]{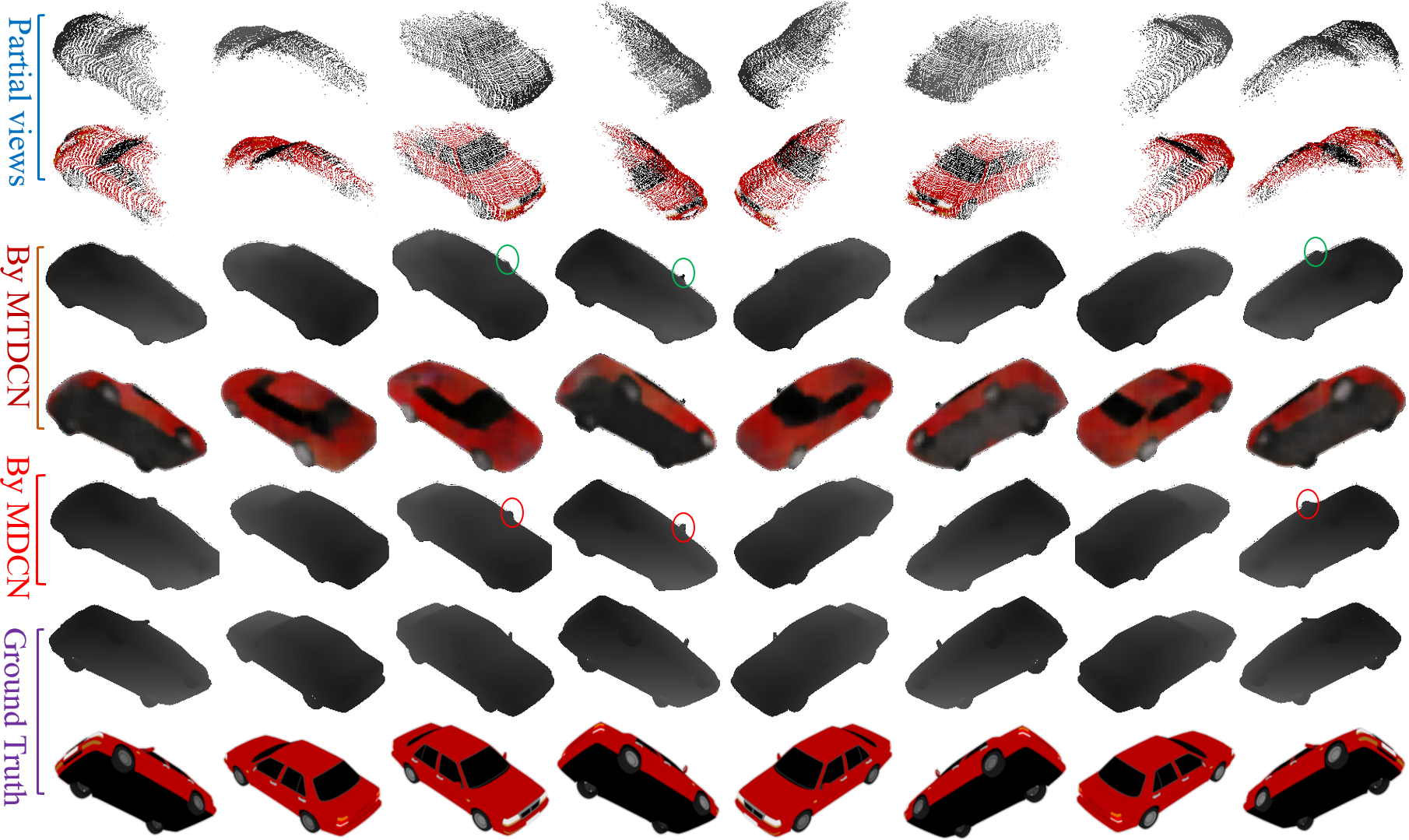}
	\end{center}
		\vspace{-0.13in}
	\caption{Completions of texture and depth maps.}
	\label{fig:views}
		\vspace{-0.1in}
\end{figure}

\noindent \textbf{Multi-view Texture-Depth Completion Net (MTDCN)}. We propose a Multi-view Texture-Depth Completion Net (MTDCN) to jointly complete texture and depth maps. MTDCN is based on a U-Net architecture. In our pipeline, we stack each pair of partial depth map $D_n$ and texture map $T_n$ into a 4-channel texture-depth map ${Q}_n=[T_n, D_n], Q_n \in \mathbb{R}^{H \times W \times 4}$, $n \in [1,8]$. MTDCN takes $Q_n$ as input, and generates completed 4-channel texture-depth maps ${Q}'_n =[T'_n, D'_n], {Q}'_n \in \mathbb{R}^{H \times W \times 4}$, where $T'_n$ and $D'_n$ are completed texture and depth map respectively. The completions of the car model are shown in Fig.~\ref{fig:views}. After fusing these views, we get a completed shape with texture $S_{dt}$ in Fig.~\ref{fig:overview}.

In contrast to the category-specific reconstruction in \cite{bird}, which samples texture from input images, thus having its performance relying on the viewpoint of the input images and the symmetry of the target objects, MTDCN can be trained to infer textures on multiple categories and does not assume objects being symmetric.

\noindent \textbf{Multi-view Depth Completion Net (MDCN)}. In our experiments, we found it very challenging to complete both depth and texture map at the same time. As an alternative we also train MDCN, which only completes partial depth maps $[D_1,\dots,D_8]$ and can generate more accurate full depth maps $[\hat{D}_1,\dots,\hat{D}_8]$. We then map the texture $[T'_1,\dots,T'_8]$ generated by MTDCN to the MDCN-generated shape $S_d$ to get a reconstructed shape with texture $S_{d+t}$ as illustrated in Fig.~\ref{fig:overview}.

Different from the multi-view completion net in \cite{mvcn}, which only completes 1-channel depth maps, MTDCN can jointly complete both texture and depth maps. It should be mentioned that there is no discriminator in MTDCN or MDCN, in contrast to \cite{mvcn}.

\subsection{Joint Fusion}

With the completed texture maps $T'=[T'_1,\dots,T'_8]$ and depth maps $D'=[D'_1,\dots,D'_8]$ by MTDCN and more accurate completed depth maps $\hat{D}=[\hat{D}_1,\dots,\hat{D}_8]$ by MDCN, we jointly fuse the depth and texture maps into a colored 3D point, as illustrated in Fig.~\ref{fig:overview}.

\noindent \textbf{Joint Fusion for MTDCN}. Given one point $p^{D'}_i = [{u}^{D'}_i, {v}^{D'}_i, {d}^D_i]$ on $D'_n$, and the aligned point $p^{T'}_i = [{u}^{T'}_i, {v}^{T'}_i, {r}^{T'}_i, {g}^{T'}_i, {b}^{T'}_i]$ on the texture map $T'_n$, where ${u}^{D'}_i={u}^{T'}_i$ and ${v}^{D'}_i={v}^{T'}_i$, the back-projected point on $S_{dt}$ is ${P'}_i=[x'_i, y'_i, z'_i, r'_i, g'_i, b'_i]$ by
\begin{equation}
{P'}_i=\Re^{-1}_s(K^{-1}p^{D'}_i-\tau_n)  \quad \forall i.
\label{eq:backproject}
\end{equation}

Note that Eq.~\ref{eq:backproject} only back-projects the depth map $D'_n$ to the coordinates of ${P'}_i$, while the texture of ${P'}_i$ is obtained from $p^{T'}_i$, where $r'_i={r}^{T'}_i, g'_i={g}^{T'}_i, b'_i={b}^{T'}_i$. We also extract a completed shape $S_d$ without texture. 

\noindent \textbf{Joint Fusion for MDCN}. We map the texture $[T'_1,\dots,T'_8]$ generated from MTDCN to the completed shape of MDCN $S_{d+t}$. The joint fusion process is similar. However, since texture and depth maps are generated separately, a valid point on a depth map may be aligned to an invalid point on the corresponding texture map, especially near edges. For such points, we take their nearest valid neighbor on the texture map. Since $S_d$ is generated by direct fusion of depth maps $[\hat{D}_1,\dots, \hat{D}_8]$, $S_{d+t}$ has the same shape as $S_d$. 

\subsection{Loss Function and Optimization}

\noindent \textbf{Training Objective}. We perform a two-stage training and train three networks: 2D-3D Net ($G1$), MTDCN ($G_2$), and MDCN ($G_3$). Given an input RGB image $X$, the generated object coordinate image is $C=G_1(X)$. The training objective of $G_1$ is  
\begin{equation}
{G_1}^*  =  \argmin_{{G_1}} ||G_1(X)-Y||_1,
\label{eq:g1}
\vspace{-0.05in}
\end{equation}
where $Y$ is the ground truth object coordinate image.

Given an partial texture-depth images ${Q}_n=[T_n, D_n]$, $n\in[1,8]$, the completed texture-depth images ${Q}'_n = G_2(Q_n)$, we get the optimal $G_2$ by 
\begin{equation}
{G_2}^*  =  \argmin_{{G_2}} ||G_2(Q_n)-Y'||_1,
\label{eq:g2}
\vspace{-0.05in}
\end{equation}
where $Y'$ is the ground truth texture-depth image.

MDCV only completes depth maps and takes 1-channel depth maps as input. Given a partial depth map $D_n$, the completed depth map $\hat{D}_n = G_3(D_n)$. $G_3$ is trained with  
\begin{equation}
{G_3}^*  =  \argmin_{{G_3}} ||G_3(D_n)-\hat{Y}||_1,
\label{eq:g3}
\vspace{-0.05in}
\end{equation}
where $\hat{Y}$ is the ground truth depth image.

\noindent \textbf{Optimization}. We use Minibatch SGD
and the Adam optimizer \cite{adam} to train all the networks. More details can be found in the supplementary material.

\section{Experiments}

We evaluate our methods (Ours-$S_{d+t}$ generated by MDCN, and Ours-$S_{dt}$ by MTDCN) on single-image 3D reconstruction and compare against state-of-the-art methods.

\noindent \textbf{Dataset and Metrics.} We train all our networks on synthetic models from ShapeNet \cite{shapenet}, and evaluate them on both ShapeNet and Pix3D \cite{pix3d}. We render depth maps, texture maps and object coordinate images for each object. More details can be found in the supplementary material. The image resolution is $256 \times 256$. We sample 100K points from each mesh object as ground truth point clouds for evaluations on ShapeNet, as in \cite{lin2018learning}. For a fair comparison, we use Chamfer Distance (CD) \cite{ref_cd} as the quantitative metric. Another popular option, Earth Mover's Distance (EMD) \cite{ref_cd}, requires that the generated point cloud has the same size as the ground truth, and its calculation is time-consuming. While EMD is often used as a metric for methods whose output is sparse and has fixed size, like 1024 or 2048 points in \cite{psgn,lmnet}, it is not suitable to evaluate our methods that generates very dense point clouds with varied number of points.

\subsection{Single Object Category}

We first evaluate our method on a single object category. Following \cite{ptn, lin2018learning}, we use the chair category from ShapeNet with the same 80\%-20\% training/test split. We compare against two methods (Tatarchenko et al. \cite{Tatarchenko2015Multiview3M} and Lin et al. \cite{lin2018learning}) that generate dense point clouds by view synthesis, as well as two voxels-based methods, Perspective Transformer Networks  (PTN) \cite{ptn} in two variants, and a baseline 3D-CNN provided in \cite{ptn}.

\begin{figure}
	\begin{center}
		\includegraphics[width=\linewidth]{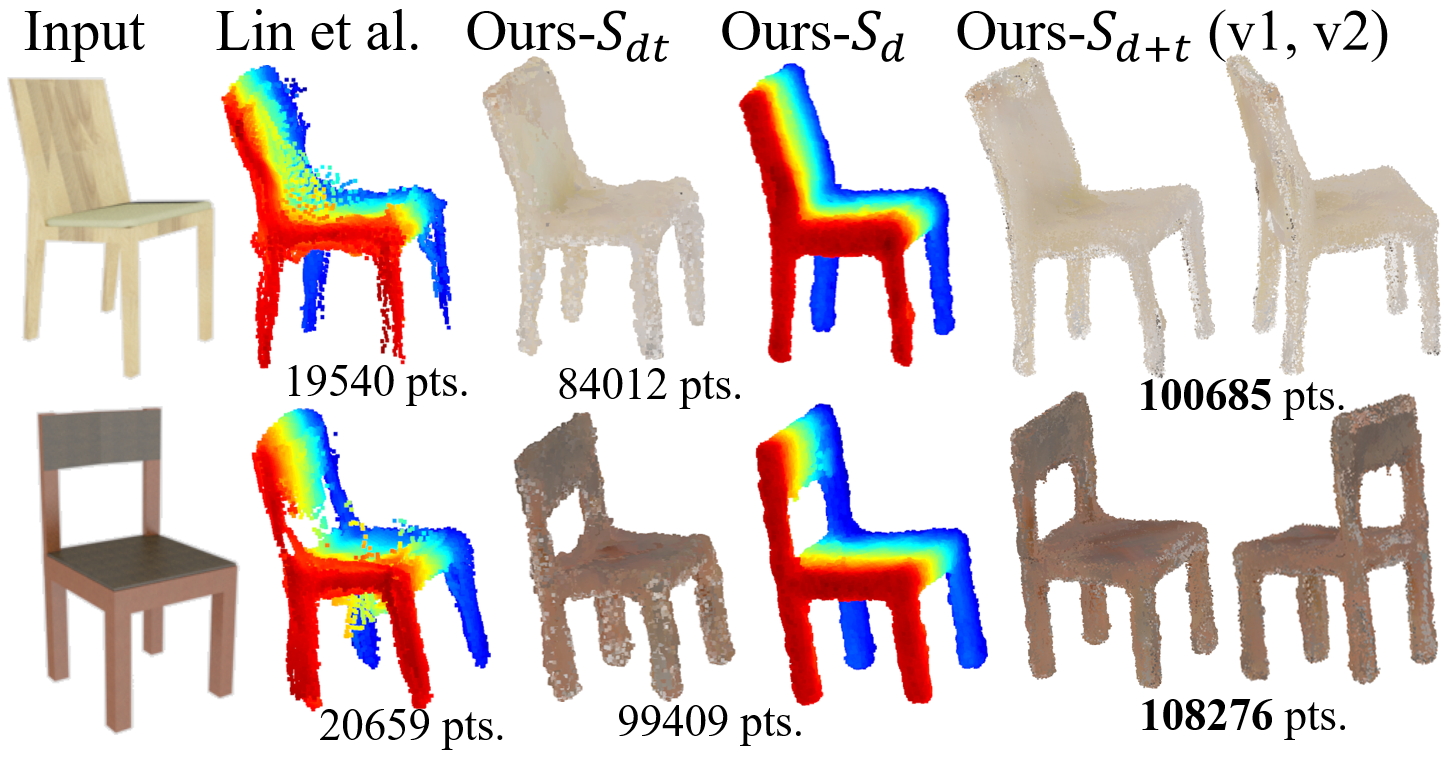}
	\end{center}
		\vspace{-0.13in}
	\caption{Reconstructions on single-category task.}
	\label{fig:chair}
	\vspace{-0.13in}
\end{figure}

The quantitative results on the test dataset are reported in
Table \ref{tab:one-category}. Test results of other approaches are referenced from \cite{lin2018learning}. Our method (Ours-$S_{d+t}$) achieves the lowest CD in this single-category task. A visual comparison with Lin's method is shown in Fig.~\ref{fig:chair}, where our generated point clouds are denser and more accurate. In addition, we also infer the textures of the generated point clouds.

\subsection{General Object Categories from ShapeNet}
We also simultaneously train our network on 13 categories (listed in Table~\ref{tab:shapenet-seen}) from ShapeNet and use the same 80\%-20\% training/test split as existing methods \cite{r2n2,lmnet}.

\noindent \textbf{Reconstruct novel objects from seen categories}. We test our method on novel objects from the 13 seen categories and compare against (a) 3D-R2N2 \cite{r2n2}, which
predicts volumeric models with recurrent networks,
and (b) PSGN \cite{psgn}, which predicts an unordered set of 1024 3D points by fully-connected layers and deconvolutional layers, and (3) 3D-LMNet which predicts point clouds by latent-embedding matching. We only compare methods that follow the same setting as 3D-R2N2, and do not include \cite{lin2018learning} which assumes fixed elevation or OptMVS \cite{mvs}. We use the pretrained models readily provided by the authors, and the results of 3D-R2N2 and PSGN are referenced from \cite{lin2018learning}. Note that we extract the surface voxels of 3D-R2N2 for evaluation.

Table~\ref{tab:shapenet-seen} shows the quantitative results. Since most methods need ICP alignment as a post-processing step to achieve finer alignment with ground truth, we list the results without and with ICP. Specially, PSGN predicts rotated point clouds, so we only list the results after ICP alignment. Ours-$S_{d+t}$ outperforms the state-of-the-art methods on most categories. Specifically, we outperform 3D-LMNet on 12 categories out of 13 without ICP, and 7 with ICP. In addition, we achieve the lowest CD in average. Different from other methods, our methods do not rely too much on ICP, and more analysis can be found in Section \ref{sec_ablation}.

\begin{figure*}[t]
	\begin{center}
		\includegraphics[width=\linewidth]{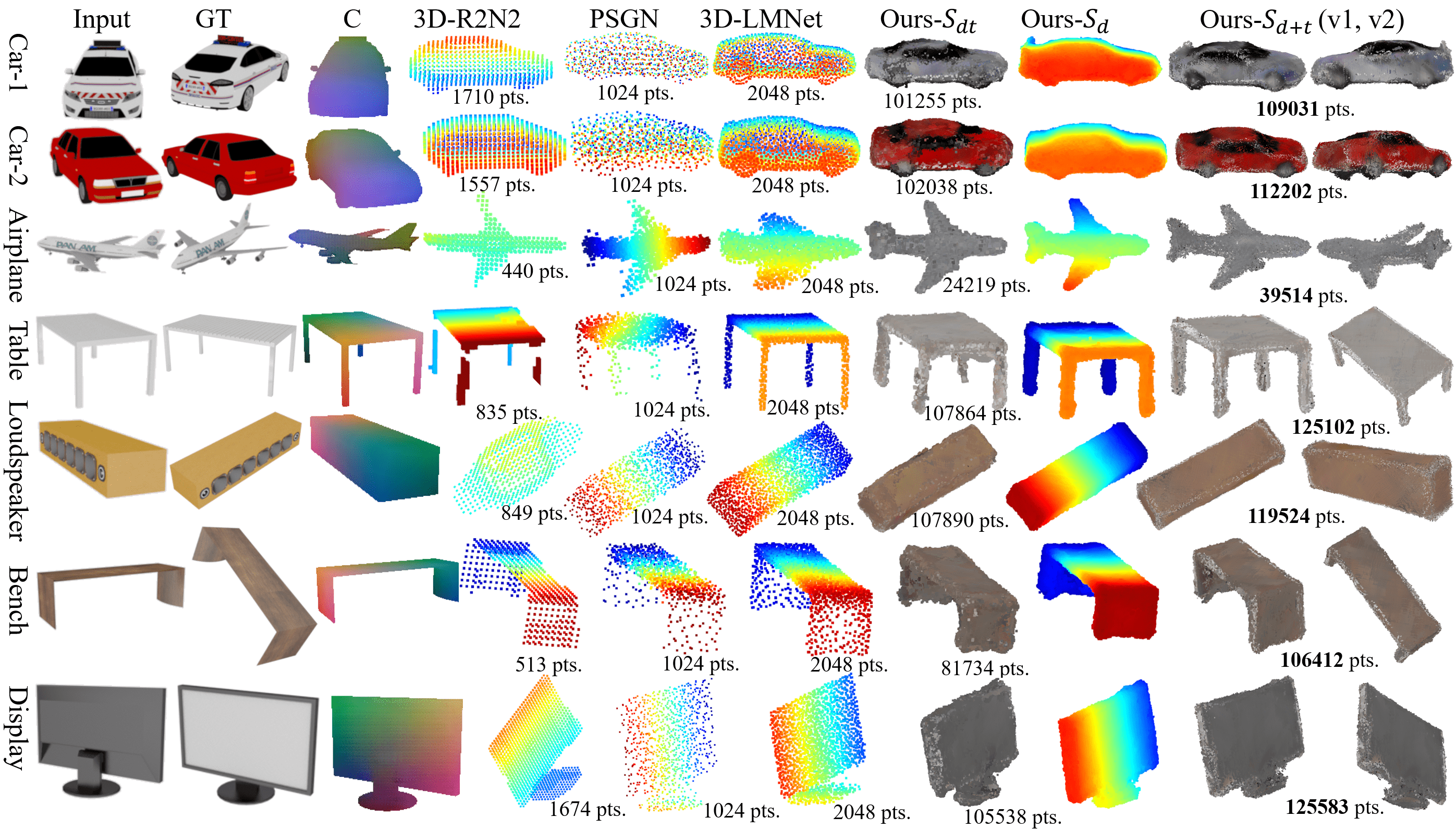}
	\end{center}
		\vspace{-0.13in}
	\caption{Reconstructions of the seen categories on ShapeNet dataset. `C' is the generated object coordinate image, and `GT' is another view of the target object.  Ours-$S_{dt}$ is generated by MTDCN, Ours-$S_{d}$ and Ours-$S_{d+t}$ are generated by MDCN. }
	\label{fig:shapenet}
	\vspace{-0.1in}
\end{figure*}

We also visualize the predictions in Fig.~\ref{fig:shapenet}. It can be seen that our method predicts more accurate shapes with higher point density. Besides 3D coordinate predictions, our methods also predict textures. We demonstrate ours-$S_{d+t}$ from two different views (v1) and (v2).

\noindent \textbf{Reconstruct objects from unseen categories}.
We also evaluate how well our models generalizes to 6 unseen categories from ShapeNet: bed, bookshelf, guitar, laptop, motorcycle, and train. The quantitative comparisons with 3D-LMNet in Table~\ref{tab:shapenet-unseen} shows a better generalization of our method. We outperform 3D-LMNet on 4 categories out of 6 before or after ICP. Qualitative completions are shown in Fig.~\ref{fig:h2}. Our methods perform reasonably well on the reconstruction of bed and guitar, while 3D-LMNet interprets the input as sofa or lamp from the seen categories respectively.

\begin{table*}[]
	\begin{minipage}{.35\linewidth}
		\begin{center}
			\begin{tabular}{|c|c|}
				\hline
				Method & CD \\ \hline \hline
				3D CNN (vol. loss only) & 4.49 \\ \hline
				PTN (proj. loss only) & 4.35 \\ \hline
				PTN (vol. \& proj. loss) & 4.43 \\ \hline
				Tatarchenko et al. & 5.40 \\ \hline
				Lin et al. & 3.53 \\ \hline
				Ours-$S_{dt}$ & 3.68 \\ \hline
				Ours-$S_{d+t}$ & \textbf{3.04}\\ \hline
			\end{tabular}
		\end{center}
			\vspace{-0.13in}
		\caption{CD on single-category task.}
		\label{tab:one-category}
		
		\begin{center}
		\vspace{-0.15in}
			\begin{tabular}{|c|c|c|}
				\hline
				Category & $P_d$ & Ours-$S_{d+t}$ \\ \hline \hline
				airplane & 10.53 & 4.19 \\ \hline
				bench & 7.85 & 3.40 \\ \hline
				cabinet & 19.07 & 4.88 \\ \hline
				car & 11.14 & 2.90 \\ \hline
				chair & 8.69 & 3.59 \\ \hline
				display & 12.43 & 4.71 \\ \hline
				lamp & 11.95 & 6.18 \\ \hline
				loudspeaker & 20.26 & 6.39 \\ \hline
				rifle & 9.47 & 5.44 \\ \hline
				sofa & 10.86 & 4.07 \\ \hline
				table & 8.83 & 3.27 \\ \hline
				telephone & 9.83 & 3.16 \\ \hline
				vessel & 9.08 & 3.79 \\ \hline
				\textbf{mean} & 10.58 & 3.91 \\ \Xhline{0.6pt} \hline \hline \Xhline{0.6pt}
				chair & 9.04 & 3.04 \\ \hline
			\end{tabular}
		\end{center}
			\vspace{-0.13in}
		\caption{Mean CD of partial shape $P_{d}$ and completed shape $S_{d+t}$ to ground truth.}
		\label{tab:contribution}
		
	\end{minipage}
	\begin{minipage}{.65\linewidth}
		\begin{center}
			\begin{tabular}{|c|c|c|c|c|c|}
				\hline
				Category & 3D-R2N2 & PSGN & 3D-LMNet & Ours-$S_{dt}$ & Ours-$S_{d+t}$ \\ \hline \hline
				airplane & (4.79) & (2.79) & 6.16 (\textbf{2.26}) & \textbf{3.70} (3.37) & 4.19 (3.66) \\ \hline
				bench & (4.93) & (3.80) & 5.79 (3.72) & 4.27 (3.83) & \textbf{3.40} (\textbf{3.10}) \\ \hline
				cabinet & (\textbf{4.04}) & (4.91) & 6.98 (4.46) & 6.77 (5.89) & \textbf{4.88} (4.50) \\ \hline
				car & (4.81) & (3.85) & 3.17 (2.91) & 2.93 (2.95) & \textbf{2.90} (\textbf{2.90}) \\ \hline
				chair & (4.93) & (4.24) & 7.08 (3.74) & 4.47 (4.12) & \textbf{3.59} (\textbf{3.22}) \\ \hline
				display & (5.04) & (4.25) & 7.89 (\textbf{3.72}) & 5.55 (4.94) & \textbf{4.71} (3.85) \\ \hline
				lamp & (13.03) & (\textbf{4.56}) & 11.36 (4.57) & 8.06 (7.13) & \textbf{6.18} (5.65) \\ \hline
				loudspeaker & (6.69) & (6.00) & 7.95 (\textbf{5.46}) & 9.53 (8.28) & \textbf{6.39} (5.74) \\ \hline
				rifle & (6.64) & (2.67) & \textbf{4.46} (\textbf{2.55}) & 5.31 (4.28) & 5.44 (4.30) \\ \hline
				sofa & (5.50) & (5.38) & 6.06 (4.44) & 4.43 (3.93) & \textbf{4.07} (\textbf{3.57}) \\ \hline
				table & (5.26) & (4.10) & 6.65 (3.84) & 4.59 (4.26) & \textbf{3.27} (\textbf{3.14}) \\ \hline
				telephone & (4.61) & (3.50) & 3.91 (3.10) & 4.98 (4.72) & \textbf{3.16} (\textbf{2.90}) \\ \hline
				vessel & (6.82) & (3.59) & 6.30 (3.81) & 4.13 (3.85) & \textbf{3.79} (\textbf{3.52}) \\ \hline
				\textbf{mean} & (5.93) & (4.13) & 6.14 (3.59) & 4.68 (4.26) & \textbf{3.91} (\textbf{3.56}) \\ \hline
			\end{tabular}
		\end{center}
			\vspace{-0.13in}
		\caption{Average CD of multiple-seen-category experiments on ShapeNet. Numbers beyond `()' are the CD before ICP, and in `()' are after ICP.}
		\label{tab:shapenet-seen}
		
		\begin{center}
			\begin{tabular}{|c|c|c|c|}
				\hline
				Category & 3D-LMNet & Ours-$S_{dt}$ & Ours-$S_{d+t}$ \\ \hline \hline
				bed & 13.56 (7.13)  & 12.82 (8.43) & \textbf{11.46} (\textbf{6.51}) \\ \hline
				bookshelf & 7.47 (\textbf{4.68}) & 8.99 (7.96) & \textbf{5.63} (4.89) \\ \hline
				guitar & 8.19 (6.40) & 7.07 (7.29) & \textbf{5.96} (\textbf{6.33}) \\ \hline
				laptop & 19.42 (\textbf{5.21}) & 9.76 (7.58) & \textbf{7.08} (5.67) \\ \hline
				motorcycle & \textbf{7.00} (5.91) & 7.32 (6.75) & 7.03 (\textbf{5.79}) \\ \hline
				train & \textbf{6.59} (4.07) & 9.16 (4.38) & 9.54 (\textbf{3.93}) \\ \hline
				\textbf{mean} & 10.37 (5.57) & 9.19 (7.06) & \textbf{7.79} (\textbf{5.52}) \\ \hline
			\end{tabular}
		\end{center}
			\vspace{-0.13in}
		\caption{Average CD of multiple-unseen-category experiments on ShapeNet.}
		\label{tab:shapenet-unseen}
		
	\end{minipage} 
\end{table*}

\begin{table*}[]
	\begin{minipage}{.75\linewidth}
		\begin{center}
			\begin{tabular}{|c|c|c|c|c|c|}
				\hline
				Category & PSGN & 3D-LMNet & OptMVS & Ours-$S_{dt}$ & Ours-$S_{d+t}$ \\ \hline \hline
				chair & (8.98) & 9.50 (\textbf{5.46}) & 8.86 (7.23) & 8.35 (7.40) & \textbf{7.28} (6.05) \\ \hline
				sofa & (7.27) & \textbf{7.82} (\textbf{6.54}) & 8.25 (8.00) & 8.54 (7.18) & 8.41 (6.83) \\ \hline
				table & (8.84) & 13.57 (\textbf{7.62}) & 9.09 (8.88) & 9.52 (9.06) & \textbf{8.53} (7.97) \\ \hline
				\textbf{mean-seen} & (8.55) & 9.73 (\textbf{6.04}) & 8.75 (7.67) & 8.54 (7.55) & \textbf{7.74} (6.53) \\ \Xhline{0.6pt} \hline \hline \Xhline{0.6pt}
				bed* & (9.23) & 13.11 (9.02) & 12.69 (9.01) & \textbf{10.91} (8.41) & 11.04 (\textbf{8.19}) \\ \hline
				bookcase* & (8.24) & 8.32 (\textbf{6.64}) & \textbf{8.10} (8.35) & 10.38 (9.72) & 8.99 (8.44) \\ \hline
				desk* & (8.40) & 11.75 (7.72) & 9.01 (8.50) & 8.64 (8.16) & \textbf{7.64} (\textbf{7.18}) \\ \hline
				misc* & (9.84) & 13.45 (11.34) & 13.82 (12.36) & 12.58 (11.03) & \textbf{11.48} (\textbf{9.30}) \\ \hline
				tool* & (11.20) & 13.64 (9.09) & 14.98 (11.27) & 13.27 (11.70) & \textbf{12.18} (\textbf{9.02}) \\ \hline
				wardrobe* & (7.84) & 9.46 (\textbf{6.96}) & \textbf{6.96} (7.26) & 9.15 (8.80) & 8.33 (8.26) \\ \hline
				\textbf{mean-unseen} & (8.81) & 11.67 (8.22) & 10.48 (8.83) & 10.19 (8.86) & \textbf{9.57} (\textbf{8.07}) \\ \hline
			\end{tabular}
		\end{center}
			\vspace{-0.15in}
		\caption{Average CD on both seen and unseen category on Pix3D dataset. All numbers are scaled by 100. `*' indicates unseen category.}
		\label{tab:pix3d-seen-unseen}
	\end{minipage}
	\begin{minipage}{.23\linewidth}
		\vspace{-0.13in}
		\begin{center}
			\includegraphics[width=.85\linewidth]{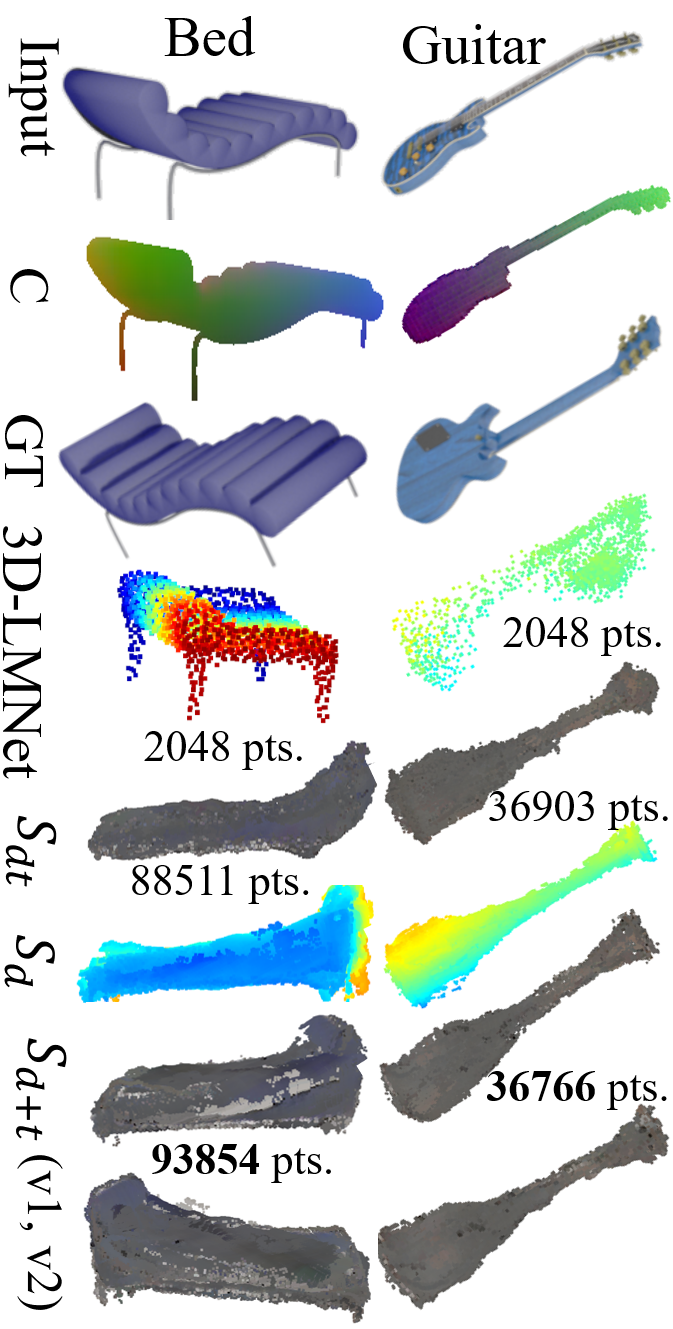}
		\end{center}
		\vspace{-0.13in}
		\captionof{figure}{Results on ShapeNet unseen category}
		\vspace{-0.18in}
		\label{fig:h2}
	\end{minipage}
\end{table*}

\subsection{Real-world Images from Pix3D}
To test the generalization of our approach to real-world images, we evaluate our trained model on the Pix3D dataset \cite{pix3d}. We compare against the state-of-the-art methods, PSGN \cite{psgn}, 3D-LMNet \cite{lmnet} and OptMVS \cite{mvs}. Following \cite{lmnet} and \cite{pix3d}, we uniformly sample 1024 points from the mesh as ground truth point cloud to calculate CD, and remove images with occlusion and truncation. We also provide the results of taking denser point cloud as ground truth in the supplementary. We have 4476 test images from seen categories, and 1048 from unseen categories.

\begin{figure*}[t]
	\begin{center}
		\includegraphics[width=\linewidth]{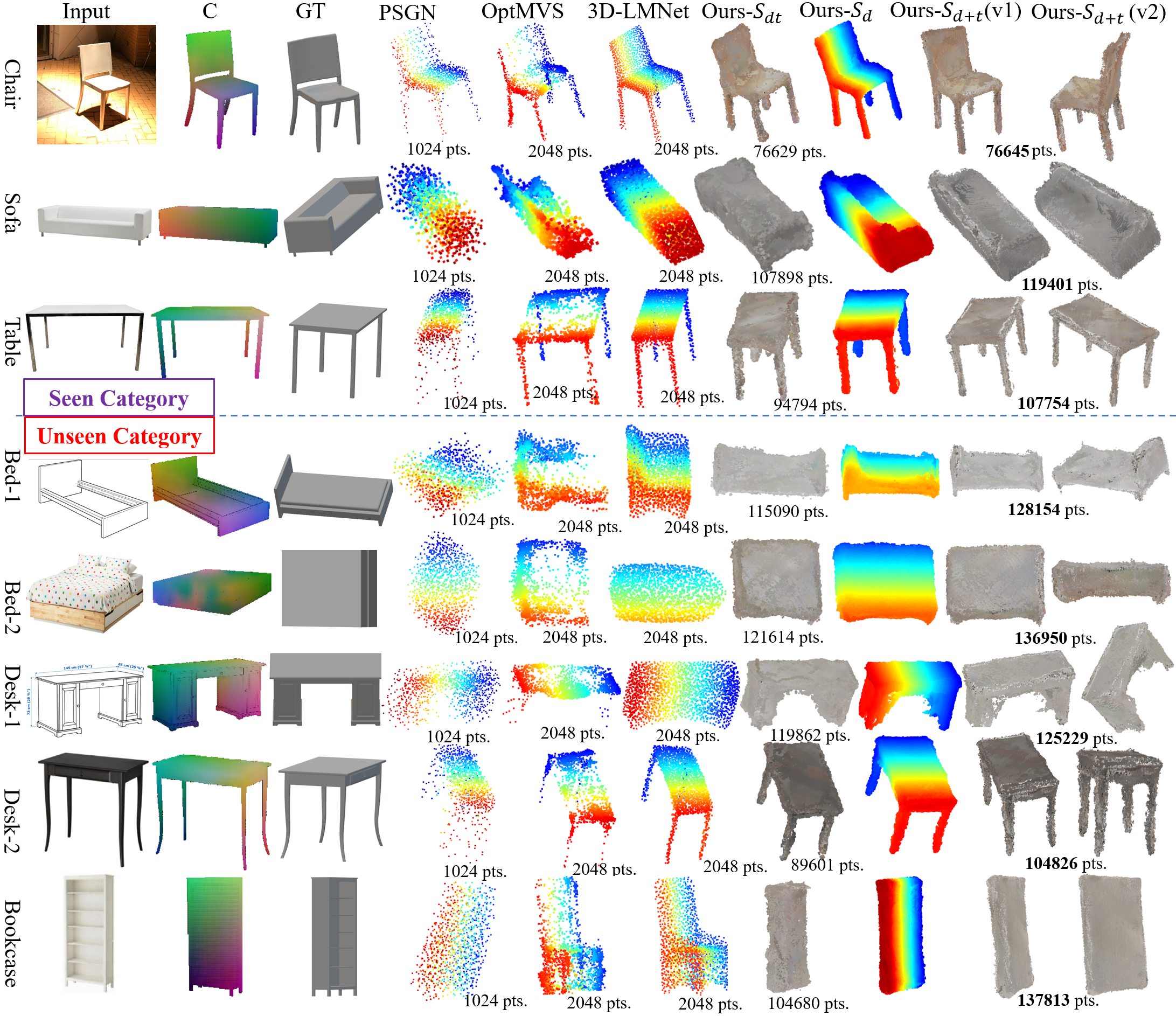}
	\end{center}
	\vspace{-0.13in}
	\caption{Reconstructions on Pix3D dataset. `C' is object coordinate image, and `GT' is  ground truth model.}
	\label{fig:pix3d}
		\vspace{-0.12in}
\end{figure*}

\noindent \textbf{Reconstruct novel objects from seen categories in Pix3D}. We test the methods on 3 seen categories (chair, sofa, table) that co-occur in the 13 training sets of ShapeNet, and the results are shown in Table ~\ref{tab:pix3d-seen-unseen}. Even on real-world data, our networks generate well aligned shapes, while other methods largely rely on ICP. Qualitative results are shown in Fig.~\ref{fig:pix3d}. Our method performs well on real images and generates denser point clouds with reasonable texture. Besides more accurate shape alignment, our method also predicts better shapes, like the aspect ratio in the `Table' example.

\noindent \textbf{Reconstruct objects from unseen categories in Pix3D}. We also test the pretrained models on 7 unseen categories (bed, bookcase, desk, misc, tool, wardrobe), and the results are shown in Table ~\ref{tab:pix3d-seen-unseen}. Our methods outperform other approaches \cite{psgn,mvs,lmnet} in mean CD with or without ICP alignment. Fig.~\ref{fig:pix3d} shows a qualitative comparison. For `Bed-1' and `Bed-2', our methods generate reasonable beds, while 3D-LMNet regards them as sofa or car-like objects. Similarly, we generate reasonable `Desk-1' and recovers the main structure of the input. 
For `Desk-2', our method estimates the aspect ratio more accurately and recovers some details of the target object, like the curved legs. For `Bookcase', ours generates a reasonable shape, while OptMVS or 3D-LMNet take it as a chair. In addition, we also successfully predict textures for unseen categories on real images.

\subsection{Ablation Study}
\label{sec_ablation}
\noindent \textbf{Contributions of each reconstruction stage to the final shape.} Considering both 2D-3D and view completion nets perform reconstruction, in Table \ref{tab:contribution}, we compare the generated partial shape $P_d$ with the completed shape Ours-$S_{d+t}$ on their CD to ground truth on the multiple-category and single-category (chair) task. For the former, the mean CD decreases from 10.58 to 3.91 after the second stage.

\noindent \textbf{Reconstruction accuracy of MTDCN and MDCN.} As shown in Tables \ref{tab:shapenet-seen},~\ref{tab:shapenet-unseen},~\ref{tab:pix3d-seen-unseen}, and Figures \ref{fig:shapenet},~\ref{fig:pix3d},~\ref{fig:chair}, MDCN generates denser point clouds with smoother surfaces, and the mean CD is lower. Fig.~\ref{fig:views} highlights that the completed maps by MDCN are more accurate than those of MTDCN.

\begin{table}[t]
		\begin{center}
			\begin{tabular}{|c|c|c|c|c|}
				\hline
				Method & S-seen & S-unseen & P-seen & P-unseen \\ \hline \hline
				3D-LMNet & 0.42 & 0.46 & 0.38  & 0.30 \\ \hline
                Ours-$S_{d+t}$ & 0.09 & 0.29 & 0.16 & 0.16 \\ \hline
			\end{tabular}
		\end{center}
			\vspace{-0.13in}
		\caption{Relative CD improvements after ICP.}
		\label{tab:icp_imp}
		\vspace{-0.18in}
\end{table}

\noindent \textbf{The impact of ICP alignment on reconstruction results.} Besides CD, pose estimation should also be evaluated in the comparisons among different reconstruction methods. We evaluate the pose estimations of 3D-LMNet and our methods by comparing the relative mean improvement of CD after ICP alignment in Table \ref{tab:icp_imp} (S: ShapeNet, P: Pix3D), which is calculated from the data in Table  \ref{tab:shapenet-seen},~\ref{tab:shapenet-unseen},~\ref{tab:pix3d-seen-unseen}. A bigger improvement means a worse alignment. Although the generated shapes of 3D-LMNet are assumed to be aligned with ground truth, its performance still relies heavily on ICP alignment. But our methods rely less on ICP, which implies that our pose estimation is more accurate. We use the same ICP implementation as 3D-LMNet \cite{lmnet}. 


\subsection{Discussion}

\label{sec:analysis}

In sum, our method predicts shape better, like pose estimation, the sizes and aspect ratio of shapes in Fig.~\ref{fig:pix3d}. We attribute this to the use of intermediate representation. The object coordinate images containing only the seen parts, are easier to infer compared to direct reconstructions from images in \cite{psgn, mvs,lmnet}. Furthermore, the predicted partial shapes also constrain the view completion net to generate aligned shapes. In addition, our method generalizes to unseen categories better than existing methods. Qualitative results in Fig.~\ref{fig:h2} and~\ref{fig:pix3d} show that our method captures more generic, class-agnostic shape priors for object reconstruction.

However, our generated texture is a little blurry since we regress pixel values, instead of predicting texture flow  \cite{bird} which predicts texture coordinates and samples pixel values directly from inputs to yield realistic textures. However, \cite{bird}'s texture prediction can only be applied on category-specific task with a good  viewpoint of the symmetric object, so it cannot be applied on multiple-category reconstruction directly. We would like to study how to combine the pixel regression methods and texture flow prediction methods together to predict realistic texture on multiple categories. 

\section{Conclusion}
We propose a two-stage reconstruction method for 3D reconstruction from single RGB images by leveraging object coordinate images as intermediate representation. Our pipeline can generate denser point clouds than previous methods and also predict textures on multiple-category reconstruction tasks. Experiments show that our method outperforms the existing methods on both seen and unseen categories on synthetic or real-world datasets. 

{\small
	\bibliographystyle{ieee_fullname}
	\bibliography{egbib}
}

\clearpage
\newpage

\addcontentsline{toc}{section}{Appendix}
\renewcommand{\thesubsection}{\Alph{subsection}}

This supplementary material provides additional experimental results and technical details for the main paper.

\subsection{Optimization}
\begin{figure}[t]
	\begin{center}
		\includegraphics[width=\linewidth]{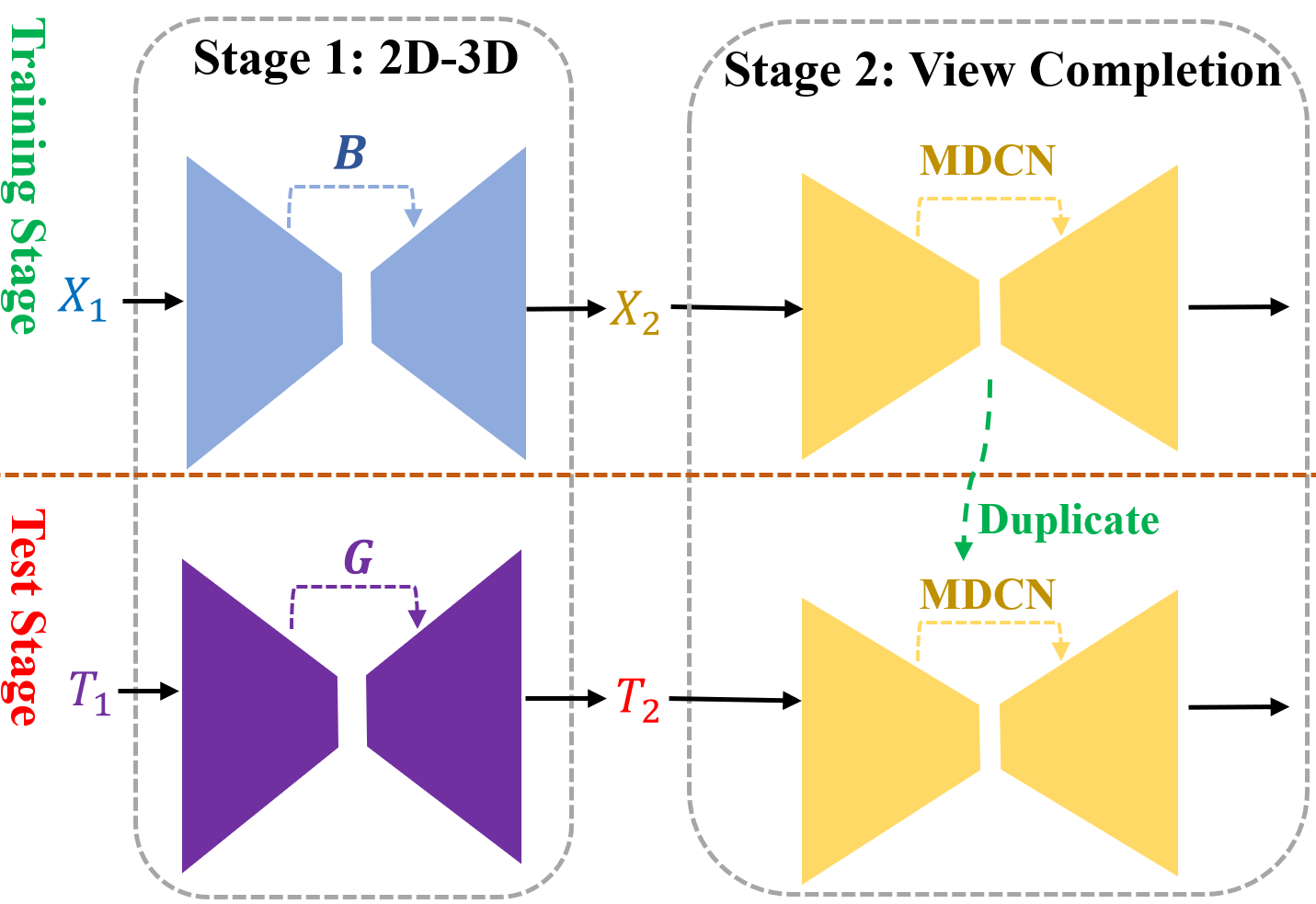}
	\end{center}
	\caption{Two-stage Training.}
	\label{fig:training}
\end{figure}

Our pipeline implements a two-stage reconstruction approach, including 2D-3D transformation by a 2D-3D net, and view completion by either the Multi-view Depth Completion Net (MDCN) or the Multi-view Texture-Depth Completion Net (MTDCN). In Fig.~\ref{fig:training} we take MDCN as an example. We implement all of our networks in PyTorch 1.2.0.

\noindent \textbf{Training 2D-3D net.} We use Minibatch SGD and the Adam optimizer \cite{adam} to train 2D-3D net, where the momentum parameters are $\beta_1=0.5$, $\beta_2=0.999$. We train 2D-3D net for 200 epochs with an initial learning rate of 0.0009, and the learning rate linearly decays after 100 epochs. The batch size is 64.




We train our networks in two stages, as shown in Fig.~\ref{fig:training}. We denote the training input data (single RGB images) of 2D-3D net as $X_1$, and test data as $T_1$. The training input data (object coordinate images) of MDCN is $X_2$, and test data is $T_2$, which correspond to the output of 2D-3D net given $X_1$ or $T_1$ as input respectively.

Let us denote $f(C)$ as the average error of $C$, like the average $L_1$ distance to ground truth. In our case, $C$ is a set of object coordinate images. In general, since $X_1$ is available during training while $T_1$ is novel input, $f(X_2)$ is smaller than $f(T_2)$ by a large margin, that is, the relative difference $ \epsilon = |f(X_2)-f(T_2)|/f(X_2)$ is large. For example, the training input $X_2$ is often less noisy than the test data $T_2$, which results in a large $\epsilon$ and limits the generalizability of MDCN. In contrast, a smaller $\epsilon$ leads to better generalizability. 


To decrease $\epsilon$, we train two 2D-3D networks separately, a `good' net ($G$) and a relatively `bad' ($B$) one such that $B$'s performance on the training set, $f(B(X_1))$, is similar to $G$'s performance on the test set, $f(G(T_1))$. In this way, $X_2=B(X_1)$ will look similar to $T_2=G(T_1)$, which improves the generalizability of MDCN. We control the number of training samples to train the two networks. $G$ is trained with 8 random views per 3D object, while $B$ is trained with only 1 for each. Note that we trained our net on both a category-specific task and a multiple-category task, hence we obtained 4 networks in total, 2 for each task.

\noindent \textbf{Training multi-view completion net.} Different from the training of 2D-3D net, we only need to train one view completion net for both MTDCN and MDCN.

Since MTDCN and MDCN have a similar network structure as 2D-3D net, we use the same optimizer setup. For MDCN, the initial learning rate is 0.0012, and the batch size is 128. For MTDCN, the learning rate is 0.0008, and the batch size is 48. Note that in our network, we concatenate all the 8 views of a 3D object into one image whose size is 2048 $\times$ 256, so that we do not need to store a shape memory or shape descriptor for each object mentioned in~\cite{mvcn}, because they are generated on the fly on one single GPU, which makes the pipeline more efficient than~\cite{mvcn}.

\subsection{More Experimental Results}

\begin{table*}[t]
	\begin{center}
		\begin{tabular}{|c|c|c|c|c|c|}
			\hline 
			Category & PSGN & 3D-LMNet & OptMVS & Ours-$S_{dt}$ & Ours-$S_{d+t}$ \\ \hline \hline
			chair & (8.36) & 8.90 (\textbf{4.72}) & 8.20 (6.54) & 7.76 (6.78) & \textbf{6.66} (5.37) \\ \hline
			sofa & (6.33) & \textbf{6.84} (\textbf{5.53}) & 7.24 (7.05) & 7.61 (6.19) & 7.47 (5.84) \\ \hline
			table & (8.07) & 12.88 (\textbf{6.79}) & 8.24 (8.06) & 8.87 (8.37) & \textbf{7.82} (7.20) \\ \hline
			\textbf{mean-seen} & (7.84) & 9.02 (\textbf{5.23}) & 7.98 (6.89) & 7.86 (6.83) & \textbf{7.03} (5.76) \\ \Xhline{0.6pt} \hline \hline \Xhline{0.6pt}
			bed* & (8.47) & 12.39 (8.24) & 11.91 (8.19) & \textbf{10.21} (7.67) & 10.31 (\textbf{7.41}) \\ \hline
			bookcase* & (7.49) & 7.49 (\textbf{5.77}) & \textbf{7.17} (7.44) & 9.54 (8.86) & 8.18 (7.61) \\ \hline
			desk* & (7.70) & 11.06 (6.98) & 8.15 (7.73) & 7.97 (7.45) & \textbf{6.91} (\textbf{6.42}) \\ \hline
			misc* & (9.36) & 12.98 (10.92) & 13.28 (11.96) & 12.10 (10.53) & \textbf{10.97} (\textbf{8.80}) \\ \hline
			tool* & (10.92) & 13.39 (8.80) & 14.69 (10.98) & 12.97 (11.39) & \textbf{11.89} (\textbf{8.71}) \\ \hline
			wardrobe* & (6.96) & 8.52 (\textbf{5.95}) & \textbf{5.89} (6.27) & 8.25 (7.89) & 7.52 (7.43) \\ \hline
			\textbf{mean-unseen} & (8.08) & 10.95 (7.45) & 9.65 (8.03) & 9.48 (8.12) & \textbf{8.85} (\textbf{7.31}) \\ \hline
		\end{tabular}
	\end{center}
	\caption{Average Chamfer Distance (CD) \cite{ref_cd} on both seen and unseen category on Pix3D  \cite{pix3d} dataset with 40K points as ground truth point cloud. All numbers are scaled by 100, and ‘*’ indicates unseen category. Numbers beyond `()' are the CD before ICP alignment \cite{icp_alg}, and in `()' are after ICP.}
	\label{tab:pix3d_dense}
\end{table*}

\begin{figure*}[t]
	\begin{center}
		\includegraphics[width=\linewidth]{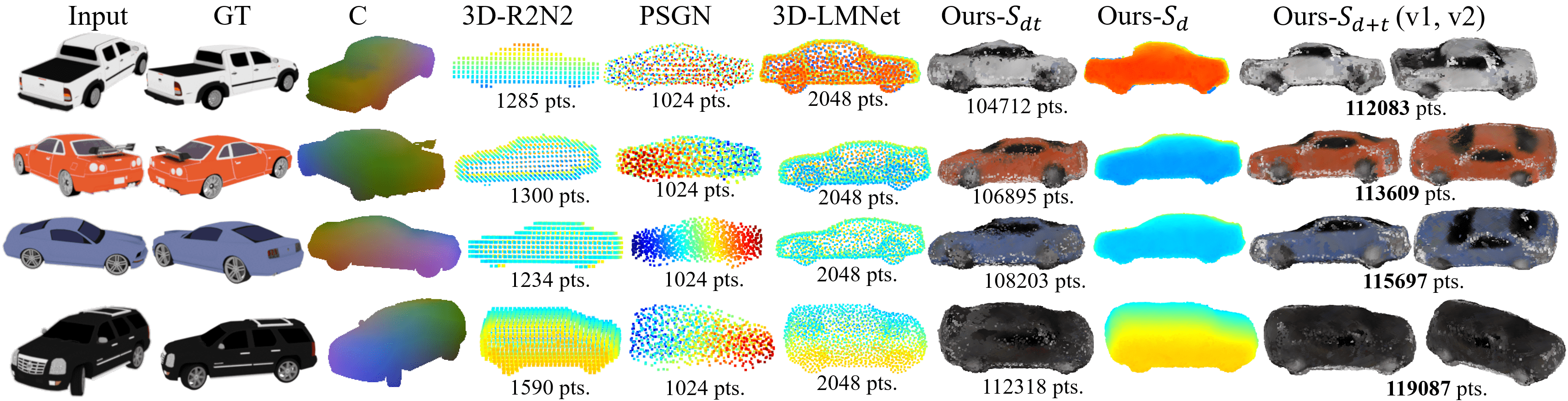}
	\end{center}
	\caption{Reconstructions of car objects on ShapeNet dataset. `C' is the generated object coordinate image, and `GT' is another view of the target object.  Ours-$S_{dt}$ is generated by MTDCN, Ours-$S_{d}$ and Ours-$S_{d+t}$ are generated by MDCN.}
	\label{fig:cars}
\end{figure*}

\noindent \textbf{Quantitative results on Pix3D.} We report the results of taking dense point clouds as ground truth in Table~\ref{tab:pix3d_dense}. Each ground truth point cloud has 40K points, different from Table \ref{tab:pix3d-seen-unseen} in the paper, which is tested with 1024 points for each ground truth point cloud. Our method is the best on both dense and sparse ground truth point clouds, compared with existing methods (e.g., PSGN \cite{psgn}, 3D-LMNet \cite{lmnet}, OptMVS \cite{mvs}).

\noindent \textbf{Qualitative results of novel car objects from ShapeNet.} Among the 13 seen categories from ShapeNet \cite{shapenet}, car objects generally have more distinct textures. In this part, we show more qualitative completions of cars in Fig.~\ref{fig:cars}, and compare against 3D-R2N2 \cite{r2n2}, PSGN \cite{psgn}, and 3D-LMNet \cite{lmnet}. We can generate denser point clouds with reasonable textures given inputs with different colors or shapes. It should be mentioned that for the first car object, our method $S_{d+t}$ generate the correct shape, while other methods fail.


\subsection{Dataset Processing}

We describe how we prepare our data for network training and testing. The dataset we use is ShapeNet \cite{shapenet}. For each model, we render 8 RGB images at random viewpoints as input, and 8 depth/texture image pairs as ground truth in MDCN training. All images have size $256\times 256$.

\noindent \textbf{Scene setup.} The camera has a fixed distance, 2.0, to the object center, which coincides with the world origin. It always looks at the origin, and has a fixed up vector $(0,1,0)$. What vary among the viewpoints is the location of the camera.




\noindent \textbf{Rendering of RGB images.} We use the Mitsuba renderer \cite{Mitsuba} to render all RGB images. 

\noindent \textbf{Rendering of depth images.} Unlike previous works \cite{ref_pcn, lin2018learning} which use a graphics engine like Blender to render depth images, ours utilizes a projection method that is similar to the Joint Projection introduced in Section 3.2 of the main paper. However, different from Joint Projection which projects partial shapes, the ground truth shape for each object is denser, which has 100K points sampled from mesh models, and the depth buffer is increased from $5 \times 5$ to $50 \times 50$ to alleviate collision effects. Because our   projection method is mainly based on matrix calculation, it renders depth maps faster than ray tracing of graphics engines.



\noindent \textbf{Rendering of object coordinate images.} Following the depth projection pipeline, we  also render object coordinate images as the ground truth to train the 2D-3D nets. First, since in our method, the object coordinate images represent the observed parts of objects, we render a depth map from the viewpoint of the input RGB image by projection method. Next, we back-project the depth map into a partial shape \{$P_i=[x_i,y_i,z_i]$\}, which can be represented by an object coordinate image, where RGB values are $[x_i,y_i,z_i]$. It should be mentioned that the input RGB image, the intermediate representation of depth map, and the object coordinate image has the same pose, which means they are aligned in pixel level.

\noindent \textbf{Fusion of depth maps.} We fuse the 8 completed depth maps into a point cloud with the Joint Fusion techniques introduced in the main paper. We also use voting algorithm to remove outliers as mentioned in \cite{mvcn}. We reproject each point of one view into the other 7 views, and if this point falls on the shape of other views, one vote will be added. The initial vote number for each point is 1, and we set a vote threshold of 5 to decide whether one point is valid or not. In addition, radius outlier removal method is used to remove noisy points that have less than 6 neighbors in a sphere of radius 0.012 around them. However, according to our experimental results, these post-processing methods have little effect on the quantitative results. For example, for single-category task (shown in Table \ref{tab:one-category} in the main paper), the Chamfer Distance decreases from 3.09 to 3.04 after these post-processing steps.


\end{document}